\documentclass[10pt,twocolumn,letterpaper]{article}
\ifdefined\anonymousreview
  \usepackage[review]{cvpr}
\else
  \usepackage[pagenumbers]{cvpr}
\fi

\usepackage{multirow}
\definecolor{cvprblue}{rgb}{0.21,0.49,0.74}
\usepackage[pagebackref,breaklinks,colorlinks,allcolors=cvprblue]{hyperref}
\usepackage[capitalize]{cleveref}

\def\confName{CVPR}
\def\confYear{2027}
\title{Progressive Pseudo-Label Optimization for Point-Supervised Change Detection}
\author{%
Hailong Ning\textsuperscript{1}, HaoWang\textsuperscript{2},
Yimeng Wang\textsuperscript{1}, Tao Lei\textsuperscript{3},\\
Nan Wang\textsuperscript{4}, Renwei Dian\textsuperscript{4},
Asoke K Nandi\textsuperscript{5}\\[0.5em]
{\normalsize\textsuperscript{1}Xi'an University of Posts and Telecommunications}\\
{\normalsize\textsuperscript{2}Dalian Martime University}\\
{\normalsize\textsuperscript{3}Shahjalal University of Science and Technology}\\
{\normalsize\textsuperscript{4}Hunan University}\\
{\normalsize\textsuperscript{5}Brunel University Uxbridge}%
}

\begin{document}
\maketitle

\begin{abstract}
Point-supervised change detection (PS-CD) aims to identify pixel-level changes between bi-temporal images using only sparsely annotated points. 
Although point annotations substantially reduce labeling costs, their limited spatial coverage often results in incomplete and noisy pseudo-labels. 
To address this issue, we propose a two-stage framework that introduces SAM2 priors into PS-CD and progressively adapts them to the target task. 
In Stage I, SAM2 generates object-aware candidate masks from point annotations on the bi-temporal images, and a bi-temporal mask selection strategy is designed to convert generic segmentation responses into more reliable change pseudo-labels. 
Subsequently, a lightweight CNN refinement module with an uncertainty-aware loss is employed to improve boundary quality and local structural consistency. 
In Stage II, we construct a teacher-student self-training framework in which the teacher is updated by exponential moving average and periodically refreshes the pseudo-labels. 
This design establishes a closed-loop optimization process that alternates between pseudo-label refinement and model re-optimization. 
We also develop a local point-interactive annotation tool that uses the bi-temporal mask selection mechanism to generate editable change masks from point prompts.
Experiments on three benchmark datasets, including WHU-CD, LEVIR-CD, and SYSU-CD, demonstrate that the proposed method outperforms previous weakly supervised approaches on most benchmarks and remains competitive with several fully supervised methods.
All code will be made publicly available upon acceptance.
\end{abstract}

\section{Introduction}

\noindent Change Detection (CD) aims to identify changes from multitemporal images of the same area and plays an important role in applications such as disaster response, urban monitoring, and ecosystem assessment.

Mainstream CD methods rely on dense pixel-level annotations for training. 
In particular, CNN-based methods~\cite{AEKAN} have shown strong capability in local feature extraction and dense prediction, while Transformer-based methods~\cite{10286068} further enhance global-context modeling and long-range dependency learning. 
More recently, foundation-model-based methods, such as ChangeCLIP~\cite{DONG202453} and SAM-based methods~\cite{11193785}, have further advanced change representation learning. 
Despite their promising performance, training and adapting these increasingly powerful models generally require substantial amounts of accurately annotated data, further intensifying the annotation bottleneck in change detection.  

\def\w{0.95\linewidth}
\def\h{1.0in}
\begin{figure}[tbp]
    \setlength{\tabcolsep}{1.0pt}
    \centering
    \small
    \begin{tabular}{@{}c@{}}
        \includegraphics[width=\w]{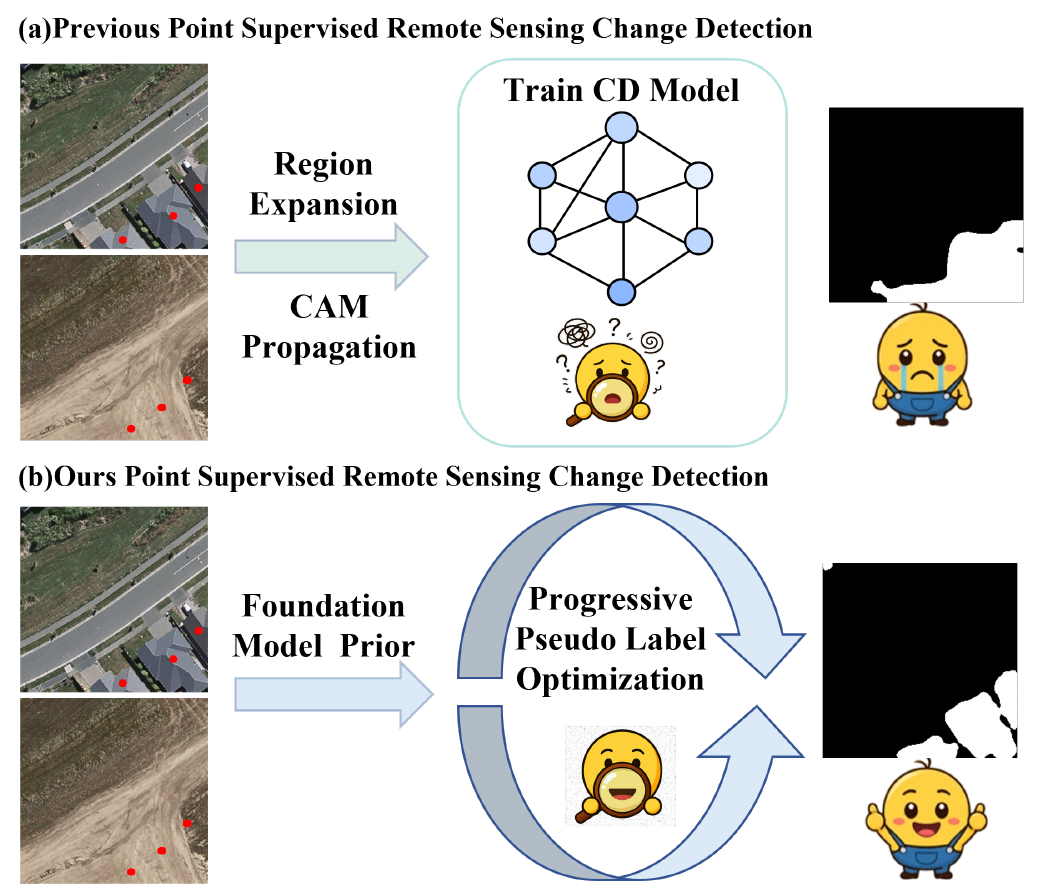}
    \end{tabular}
    \caption{The Motivation of progressive pseudo-label refinement.}
    \label{fig:intro}
\end{figure}

Point supervision~\cite{subhani2026resam} provides a representative weakly supervised alternative by replacing pixel-level masks with a small number of annotated change points. 
Although point annotations provide explicit localization cues at a substantially lower cost, they convey little information about object extent, boundary structure, or background regions. 
Therefore, directly training a dense change detector from such sparse annotations is highly challenging. 
A common solution is to expand point annotations into pixel-level pseudo-labels and use them as surrogate supervision. 
Consequently, the quality of pseudo-labels becomes a key factor determining the performance of point-supervised change detection.
For practical annotation, these expanded regions must also be accessible for visual inspection and correction, motivating an interactive workflow that turns sparse clicks into editable change masks.

As illustrated in Figure~\ref{fig:intro}, existing methods mainly generate pseudo-labels through region expansion~\cite{fang2023point} or class activation maps (CAMs) propagation~\cite{10310006}. 
Such strategies typically rely on local appearance similarity or coarse discriminative responses, which may result in incomplete regions, inaccurate boundaries, and false activations. 
Moreover, pseudo-labels are typically generated only once and subsequently treated as fixed supervision, allowing early errors to propagate throughout training. 
Foundation models provide rich object-level segmentation priors and offer a promising way to improve pseudo-label generation. 
Nevertheless, their direct application to change detection remains nontrivial because their predictions are usually produced independently for single-temporal images and are not specifically adapted to the appearance variations encountered in change detection. 
This raises two central questions: \textit{how can foundation-model priors be converted into reliable bi-temporal change pseudo-labels, and how can these pseudo-labels be progressively improved during training rather than remaining fixed?}

To address these issues, we propose a two-stage framework for point-supervised change detection. In Stage I, SAM2~\cite{ravi2024sam2} is prompted with sparse point annotations on the bi-temporal images to generate object-aware candidate masks. Then, a \textit{Bi-Temporal Mask Selection} mechanism is proposed to rank the candidates by jointly considering SAM confidence,
change consistency, and point coverage to convert generic single-temporal segmentation responses into initial change pseudo-labels. Since SAM2 priors may still contain imprecise boundaries and local noise, we further introduce a lightweight \textit{CNN-based Pseudo-Label Refinement} module with an uncertainty-aware loss to improve boundary quality and local structural consistency while suppressing unreliable supervision. In Stage II, we construct an EMA-based \textit{Teacher-Student Self-Training} framework. 
Instead of treating the pseudo-labels generated in Stage I as fixed supervision, the teacher model is updated using the exponential moving average (EMA) of the student parameters and periodically refreshes the pseudo-labels with its predictions. This design enables the model to continuously improve its supervision during training, thereby forming a closed-loop learning process of model optimization, pseudo-label updating, and re-optimization.

We further implement the candidate generation and bi-temporal mask selection mechanism as a local point-interactive annotation tool (Figure~\ref{fig:software}). It uses pretrained SAM2 to generate change masks from user clicks, with support for subsequent correction and export, without additional task-specific training.
\def\w{\linewidth}
\def\h{1.0in}
\begin{figure}[tbp]
    \setlength{\tabcolsep}{1.0pt}
    \centering
    \small
    \begin{tabular}{@{}c@{}}
        \includegraphics[width=\w]{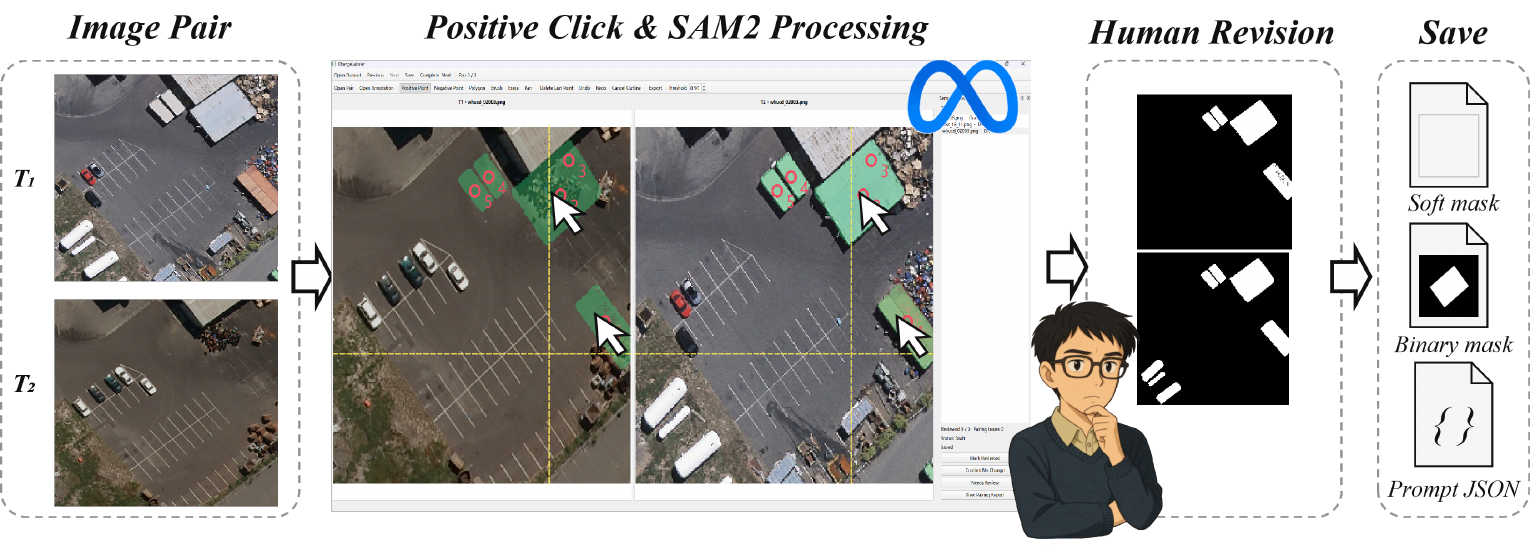}
    \end{tabular}
    \caption{Workflow of the interactive change annotation tool. Positive clicks on a registered image pair trigger SAM2 candidate generation and bi-temporal mask selection. Users can inspect the resulting change masks, optionally revise them, and export masks and prompt records.}
    \label{fig:software}
\end{figure}
The main contributions of this study are summarized as follows:
\begin{itemize}
    \item We introduce a progressive pseudo-label optimization paradigm for point-supervised change detection, which jointly evolves pseudo-label quality and detector capability through iterative optimization.
    \item We propose a SAM2-guided bi-temporal mask selection strategy that transforms single-temporal foundation-model responses into change-aware soft pseudo-labels.
    \item We introduce an uncertainty-aware refinement and EMA-based self-training framework that improves pseudo-label boundary quality and periodically refreshes supervision during training.
    \item We develop a local point-interactive change annotation tool that integrates bi-temporal mask selection with user correction and traceable mask export.
\end{itemize}

\section{Related Work}
\noindent \textbf{Weakly Supervised Change Detection in Remote Sensing.}
Remote sensing change detection has achieved remarkable progress with the development of deep neural networks. 
However, most existing methods~\cite{chen2021a} rely on dense pixel-level annotations, which are costly and labor-intensive to obtain for large-scale remote sensing imagery. 
To reduce annotation burden, recent studies~\cite{fang2023point, jiang2025barnet, zhao2025transwcd} have explored various weakly supervised settings, including image-level, point-level, and coarse region-level supervision. 
Among them, point-level supervision is particularly attractive because it provides explicit localization cues while requiring substantially less annotation effort than dense change masks.
Most weakly supervised change detection methods follow a pipeline similar to weakly supervised semantic segmentation, in which limited annotations are first converted into pseudo-labels and then used to train a dense predictor. 
For example, CS-WSCD~\cite{10310006} employs CAMs to coarsely localize changed regions and subsequently introduces SAM to refine ambiguous areas. 
FCD-GAN~\cite{wu2022fullyconvolutionalchangedetection} unifies unsupervised, weakly supervised, and regionally supervised change detection within a common framework, although its adversarial training may suffer from optimization instability. 
CARGNet~\cite{fang2023point} expands sparse point annotations into change regions through consistency-aligned regional growth, while BARNet~\cite{jiang2025barnet} and TransWCD~\cite{zhao2025transwcd} further improve weakly supervised change localization with stronger feature modeling. 
Nevertheless, these methods remain highly dependent on the quality of their pseudo-labels, which are commonly generated by heuristic expansion or coarse localization strategies. Consequently, how to reliably recover complete and accurate change regions from weak supervision, particularly from point-level supervision, remains largely underexplored.

\noindent \textbf{Pseudo-Label Learning and Self-Training.}
Pseudo-label learning~\cite{Kage_2026} is a fundamental technique in weakly supervised dense prediction, since the final performance largely depends on whether limited annotations can be transformed into reliable pixel-level supervision. 
Similar strategies have also been successfully applied to camouflaged object detection~\cite{he2025scalersamenhancedcollaborativelearning} and medical object detection~\cite{meyer2026dexterweaklysemisupervisedobject}, where progressively improved pseudo-labels provide effective supervision under limited annotations.
In remote sensing change detection, however, many existing methods still adopt a one-shot pipeline~\cite{10310006}, in which pseudo-labels are generated once and subsequently treated as fixed supervision throughout training. 
Such a strategy may propagate early pseudo-label errors and limits the model's ability to progressively adapt to the target task.
In contrast, self-training~\cite{10839471} provides a natural way to iteratively improve pseudo-labels through model feedback. 
Motivated by this principle, our method does not treat SAM-generated pseudo-labels as fixed supervision and progressively updates pseudo-labels within a teacher-student framework.

\def\w{1.0\linewidth}
\def\h{1.0in}
\begin{figure*}[!t]
    \setlength{\tabcolsep}{1.0pt}
    \centering
    \small
    \begin{tabular}{@{}c@{}}
        \includegraphics[width=\w]{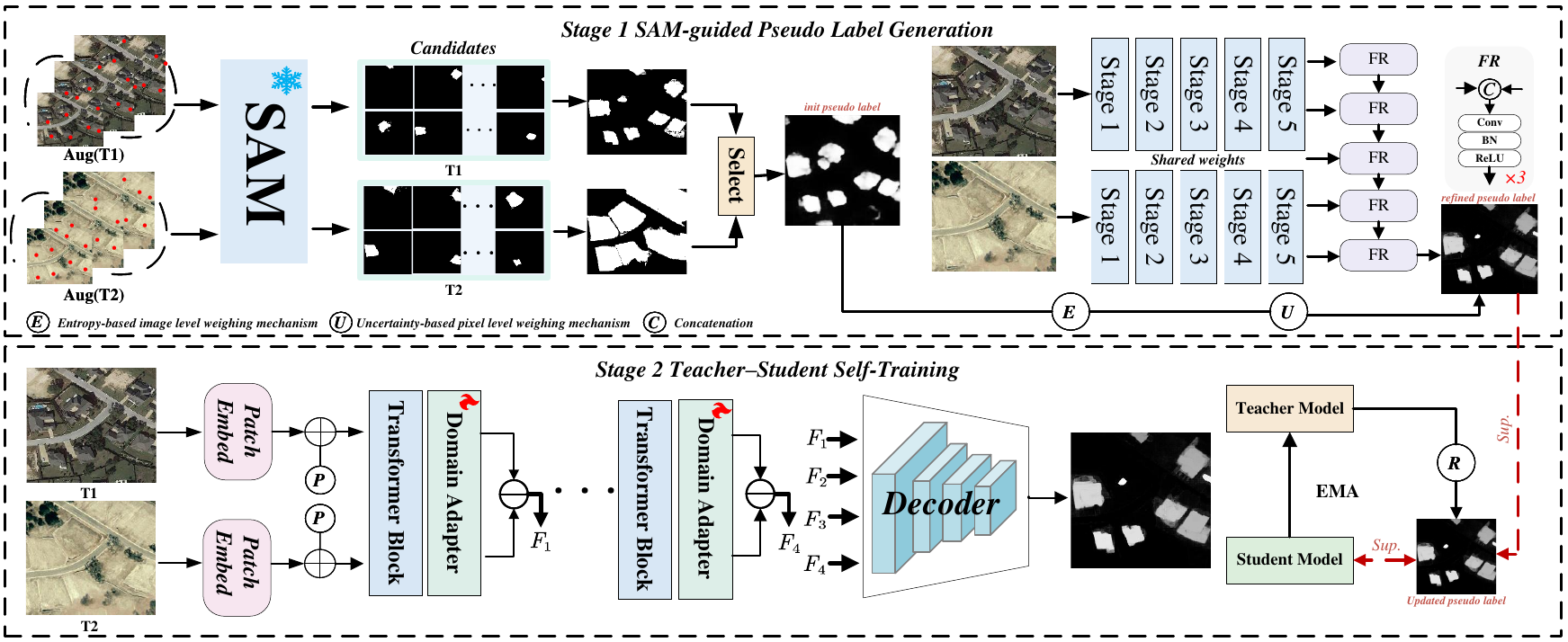}
    \end{tabular}
    \caption{Overview of the proposed two-stage point-supervised change detection framework. Stage I generates and refines soft pseudo-labels from bi-temporal point supervision using SAM2, mask selection, and CNN-based refinement. Stage II further improves change detection performance through a teacher-student self-training framework with EMA update and periodic pseudo-label refresh.}
    \label{fig:Framework}
\end{figure*}

\noindent \textbf{Foundation Models for Change Detection in Remote Sensing.}
The Segment Anything Model (SAM)~\cite{kirillov2023segment} exhibits strong zero-shot segmentation capability and provides a new opportunity for remote sensing change detection. 
However, SAM is pretrained on natural single-temporal images, and its direct application to remote sensing scenarios is limited by both the domain gap and the inherently bi-temporal nature of the task. 
To adapt foundation models to change detection, SAM-CD~\cite{Ding_2024} introduces a convolutional adapter and a task-agnostic semantic learning branch to align bi-temporal representations. 
Subsequent studies~\cite{rs17030369} further explored different adaptation strategies. 
Although these methods demonstrate the potential of foundation models for change detection, they are mainly developed under fully supervised settings. 
How to exploit SAM/SAM2 priors to construct reliable supervision under weak annotations remains insufficiently studied.
Translating point-prompted object segmentation into interactive change annotation additionally requires selecting regions according to bi-temporal change cues rather than single-image objectness alone.

\section{Methodology}
Figure~\ref{fig:Framework} illustrates the overall framework of the proposed method, which consists of two stages under point-level supervision. 
In Stage I, SAM2 generates candidate masks for the bi-temporal images, and a bi-temporal mask selection strategy is used to rank and fuse them into initial soft pseudo-labels. These pseudo-labels are further refined by a CNN-based change detector with uncertainty-aware supervision to improve boundary quality and local structural consistency.
In Stage II, the refined pseudo-labels are used to fine-tune a SAM2 model with an adapter in a teacher-student self-training framework. The teacher network is updated by exponential moving average (EMA) and periodically refreshes the pseudo-labels, forming a closed-loop optimization process for progressive pseudo-label refinement and task-specific adaptation.
The candidate generation and selection mechanism also supports the interactive tool described in Section~\ref{sec:annotation_tool}.

\subsection{Stage I: pseudo-label Generation and Refinement}
Weak point-level supervision provides only sparse localization cues, making it difficult to recover complete object regions. 
Directly optimizing with such sparse annotations often leads to incomplete masks and noisy supervision. 
To address this issue, we first construct initial pseudo-labels by leveraging the strong segmentation prior of the foundation model SAM2. 
Although the foundation model exhibits strong zero-shot segmentation capability, the domain gap between its pretraining data and the target change detection images often leads to semantically inaccurate predictions and imprecise object boundaries, making its pseudo-labels unreliable for direct supervision.
To address this limitation, we incorporate a CNN-based ResNet branch as a complementary feature extractor.

\noindent \textbf{Step 1: SAM2-based Candidate Mask Generation.}

Given a pair of bi-temporal images \(I^{t1}\) and \(I^{t2}\), 
we first convert the weak change annotations into a set of positive point prompts \(\{p_k\}_{k=1}^{K}\), where each point corresponds to one connected change region. 
For each prompt point \(p_k\), SAM2 is independently applied to the two temporal images to generate multiple candidate masks. The generated masks at the two timestamps are denoted as the candidate mask sets \(\mathcal{M}_{k}^{t1}\) and \(\mathcal{M}_{k}^{t2}\), respectively, where each set contains \(N\) candidate masks corresponding to the same prompt point.

\noindent \textbf{Step 2: Bi-Temporal Mask Selection.}

Since SAM2 may generate noisy or over-expanded masks, we rank candidate masks using a weighted score based on three criteria: SAM2 confidence \(s_{k,n}^{sam}\), change consistency \(s_{k,n}^{chg}\), and point coverage \(s_{k,n}^{pt}\). 
The change consistency is measured by a change response map \(D\), which integrates grayscale intensity, Sobel-gradient, and high-frequency differences between bi-temporal images:
\begin{align}
    D=\mathrm{Norm}(D_{int}+D_{grad}+D_{hf}),\\
    s_{k,n}^{chg}=
    \frac{\sum_{x\in\Omega}M_{k,n}(x)D(x)}
    {\sum_{x\in\Omega}M_{k,n}(x)+\epsilon},\\
    s_{k,n}^{pt}=
    \frac{1}{|\mathcal{P}_k|}
    \sum_{p\in\mathcal{P}_k}\mathbb{I}[M_{k,n}(p)>0],
\end{align}
where \(\mathrm{Norm}(\cdot)\) denotes min-max normalization, \(\mathcal{P}_k\) represents the point prompts of the \(k\)-th component, and \(\mathbb{I}(\cdot)\) is the indicator function.
The final score and pseudo-label generation process are formulated as
\begin{align}
s_{k,n} &= \lambda_1s_{k,n}^{sam}
+\lambda_2s_{k,n}^{chg}
+\lambda_3s_{k,n}^{pt}, \nonumber\\
\tilde{Y}(x) &= \max_k(s_kM_k(x)),
\end{align}
where \(\lambda_1,\lambda_2,\lambda_3\) balance the three criteria. 
After ranking candidate masks according to \(s_{k,n}\), the selected masks are fused into component-level soft masks to generate the initial pseudo label \(\tilde{Y}\).

\noindent \textbf{Step 3: CNN-based Pseudo-Label Refinement.}
SAM2 outputs may still miss fine boundaries and local texture details. 
Therefore, we further train a CNN-based change detector using the generated soft pseudo-labels as supervision. 
Specifically, we adopt a siamese encoder-decoder architecture, where a shared ResNet-18 backbone extracts multi-scale features from the two temporal images, 
and the absolute feature differences are progressively decoded to produce a dense change probability map. This can be formulated as follows:
\begin{align}
    \left\{ F_l^{t1} \right\}_{l=1}^{L} = E(I^{t1}), \qquad
    \left\{ F_l^{t2} \right\}_{l=1}^{L} = E(I^{t2}),\\
    D_l = \left| F_l^{t1} - F_l^{t2} \right|, \qquad l=1,2,3,4\\
    P = Decoder\left(D_1, D_2,D_3 , D_4\right),
\end{align}
Since the pseudo-labels generated in the previous stage are inevitably noisy, directly treating all pixels equally may introduce unreliable supervision.
Inspired by uncertainty-aware pseudo-label learning, we assign both image-level and pixel-level confidence weights to the soft pseudo-labels, so that reliable regions contribute more to optimization while ambiguous regions are suppressed.
\begin{align}
w(x) &= \left(1-E(\tilde{Y})\right)\left(2\tilde{Y}(x)-1\right)^2,\\
E(\tilde{Y}) &= \frac{1}{|\Omega|}\sum_{x\in\Omega} H\!\left(\tilde{Y}(x)\right),\\
\mathcal{L}_{ref} &= \mathcal{L}_{bce}^{w}(P,\tilde{Y}) + \mathcal{L}_{dice}^{w}(P,\tilde{Y}),
\end{align}
where $H(\cdot)$ denotes the binary entropy function, \(\mathcal{L}_{bce}^{w}\) and \(\mathcal{L}_{dice}^{w}\) denote the weighted binary cross-entropy loss and weighted Dice loss under the uncertainty-aware weight \(w(x)\), respectively.

\def\w{0.95\linewidth}
\def\h{1.0in}
\begin{figure}[tbp]
    \setlength{\tabcolsep}{1.0pt}
    \centering
    \small
    \begin{tabular}{@{}c@{}}
        \includegraphics[width=\w]{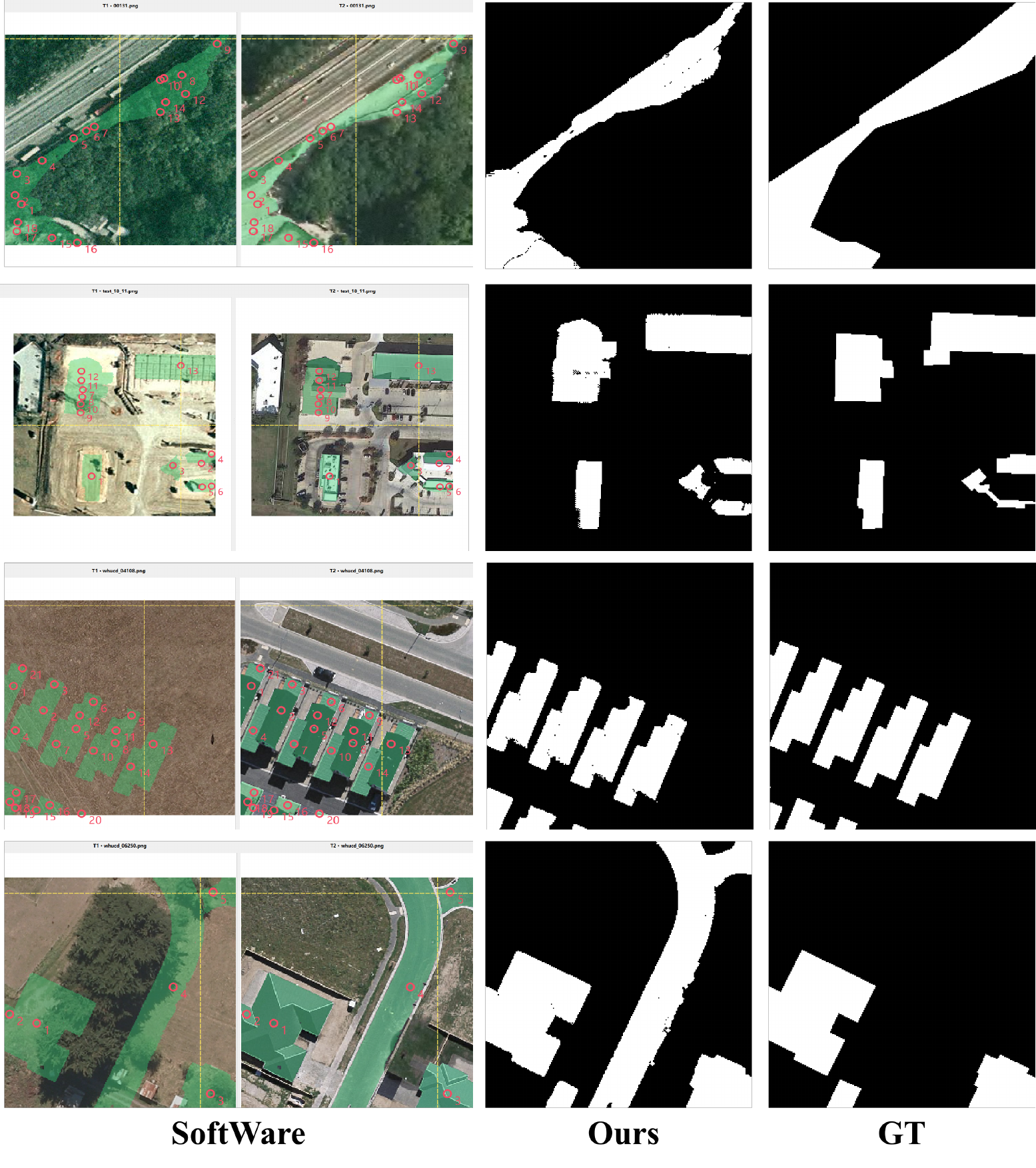}
    \end{tabular}
    \par\smallskip
    \caption{Positive points only; no manual correction. Examples of software-generated change masks. Left: paired-image annotation interface with point prompts and mask overlays. Middle: generated binary masks. Right: ground-truth masks.}
    \label{fig:cmp-software}
\end{figure}

\def\w{1.0\linewidth}
\def\h{1.0in}
\begin{figure*}[!t]
    \setlength{\tabcolsep}{1.0pt}
    \centering
    \small
    \begin{tabular}{@{}c@{}}
        \includegraphics[width=\textwidth]{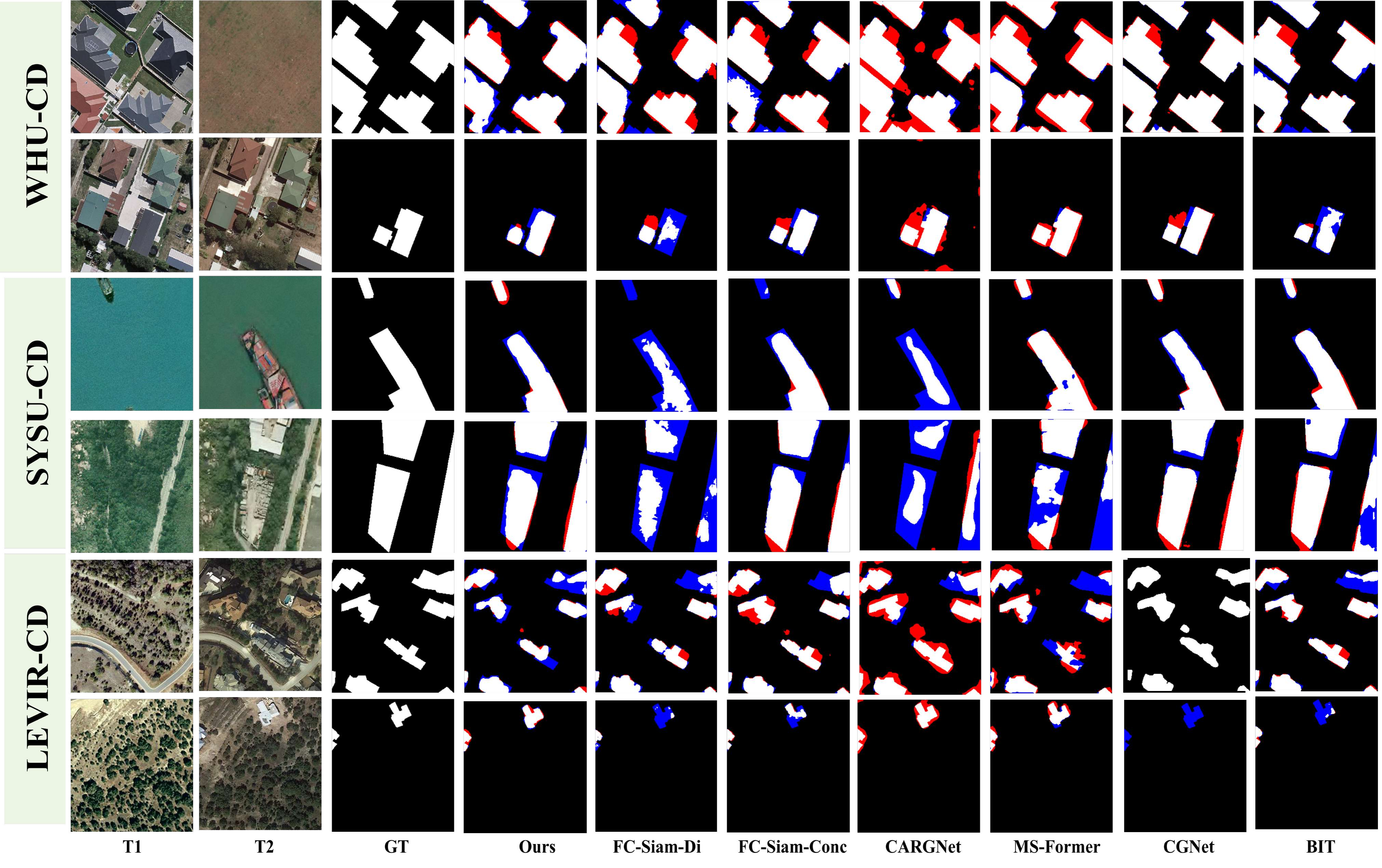}
    \end{tabular}
    \caption{Qualitative comparison of detection results on Change Detection datasets.}
    \label{fig:vis-Compare}
\end{figure*}

\subsection{Stage II: Teacher–Student Self-Training}
Different from existing one-way pipelines that use SAM2 only for initial pseudo-label generation, our goal is to establish a closed-loop optimization process, where the SAM-initialized pseudo-labels continuously interact with the task-specific change detector during training. We introduce a teacher-student self-training framework.
Given the pseudo-labels generated in Stage I, the student network is first optimized to learn bi-temporal change representations:
\begin{align}
P_s=\Psi_s\!(I^{t1},I^{t2};\theta_s),\\
P_t=\Psi_t\!\left(I^{t1},I^{t2};\theta_t\right),\\
\theta_t\leftarrow\alpha\theta_t+(1-\alpha)\theta_s,
\end{align}
where \(\Psi_s(\cdot)\) denotes the student network, \(\Psi_t(\cdot)\) denotes the teacher network, and \(\theta_s\) and \(\theta_t\) are their corresponding parameters.
The teacher network shares the same architecture as the student network, but is updated as the exponential moving average (EMA) of the student, where \(\alpha\) is the EMA decay factor. 
Such a design enables the teacher to accumulate the task-adaptive knowledge learned by the student and to provide more stable guidance during training.
The updated teacher produces more reliable predictions to refine the pseudo-labels:
\begin{align}
\tilde{Y}^{r+1}=\mu\tilde{Y}^{r}+(1-\mu)P_t,\\
\mathcal{L}_{stage2}=\mathcal{L}_{u}(P_s,\tilde{Y}^{r+1}).
\end{align}
where \(\tilde{Y}^{r}\) and \(\tilde{Y}^{r+1}\) denote the pseudo-labels before and after the \(r\)-th refresh step, respectively, and \(\mu\) is the momentum coefficient for pseudo-label updating. The refreshed pseudo-labels are then used to supervise the student through the uncertainty-aware loss \(\mathcal{L}_{u}(\cdot)\).

\subsection{Interactive Change Annotation Tool}
\label{sec:annotation_tool}
We implement the candidate generation and bi-temporal mask selection mechanism from Stage I as a local annotation tool. As shown in Figure~\ref{fig:software}, registered RGB image pairs are displayed in synchronized canvases. Users click changed regions, and pretrained SAM2 generates candidates at both timestamps; change-guided selection produces a change mask for inspection. Additional positive or negative prompts and local drawing or erasing allow subsequent correction. The tool exports soft and binary masks together with prompt coordinates, thresholds, and model metadata. This inference workflow requires no additional task-specific training. Figure~\ref{fig:cmp-software} shows outputs obtained using positive prompts only, without manual correction.

\section{Experiment}

\begin{table*}[!ht]
    \centering
    \scriptsize
    \setlength{\tabcolsep}{2.2pt}
    \renewcommand{\arraystretch}{1.1}
    \resizebox{\textwidth}{!}{
    \begin{tabular}{@{}l
        ccccc
        ccccc
        ccccc@{}}
        \toprule

        \multirow{2}{*}{\textbf{Method}}
        & \multicolumn{5}{c}{\textbf{LEVIR-CD}}
        & \multicolumn{5}{c}{\textbf{WHU-CD}}
        & \multicolumn{5}{c}{\textbf{SYSU-CD}} \\

        \cmidrule(r){2-6}
        \cmidrule(r){7-11}
        \cmidrule(r){12-16}

        & P & R & F1 & IoU & OA
        & P & R & F1 & IoU & OA
        & P & R & F1 & IoU & OA \\

        \midrule

        \multicolumn{16}{c}{\textit{Fully-supervised Methods}} \\
        \midrule

        FC-Siam-Conc~\cite{8451652}
        & 90.52 & 84.59 & 87.45 & 77.70 & 98.76
        & 90.34 & 81.04 & 85.44 & 74.58 & 98.90
        & 82.22 & 70.74 & 76.05 & 61.36 & 89.49 \\

        FC-Siam-Diff~\cite{8451652}
        & 91.66 & 81.65 & 86.36 & 76.00 & 98.69
        & 88.87 & 77.59 & 82.85 & 70.72 & 98.73
        & 81.76 & 38.95 & 52.77 & 35.84 & 83.55 \\

        BIT~\cite{chen2021a}
        & 91.33 & 88.31 & 89.79 & 81.48 & 98.98
        & 86.80 & 88.72 & 87.75 & 78.17 & 99.02
        & 80.39 & 76.28 & 78.28 & 64.31 & 90.02 \\

        SNUNet~\cite{9355573}
        & 91.53 & 88.30 & 89.89 & 81.64 & 98.99
        & 86.64 & 90.64 & 88.51 & 79.39 & 99.07
        & 81.12 & 71.16 & 75.81 & 61.04 & 89.29 \\
        
        ChangeFormer~\cite{9883686}
        & 92.05 & 88.80 & 90.40 & 82.48 & 99.04
        & 87.64 & 76.11 & 81.47 & 68.73 & 98.63
        & 79.10 & 72.64 & 75.73 & 60.94 & 89.02 \\

        HCGMNet~\cite{han2023hcgmnethierarchicalchangeguiding}
        & 93.55 & 89.65 & 91.56 & 84.43 & 99.16
        & 91.81 & 93.75 & 92.77 & 86.51 & 99.42
        & 83.93 & 75.73 & 79.62 & 66.14 & 90.86 \\
        
        CGNet~\cite{10093022}
        & 93.05 & 90.23 & 91.62 & 84.53 & 99.16
        & 92.57 & 93.81 & 93.18 & 87.24 & 99.46
        & 86.37 & 74.37 & 79.92 & 66.55 & 91.19 \\

        SAM-CD~\cite{Ding_2024}
        & 95.87 & 95.14 & 95.50 & -- & 99.14
        & 97.97 & 97.20 & 97.58 & -- & 99.60
        & 85.56 & 73.22 & 78.91 & 65.17 & 90.77 \\

        ChangeCLIP~\cite{DONG202453}
        & 93.68 & 89.04 & 91.30 & 83.99 & 99.14
        & 96.02 & 93.58 & 94.78 & 90.08 & 95.52
        & 87.16 & 79.80 & 83.82 & 71.41 & 92.46 \\

        \midrule

        \multicolumn{16}{c}{\textit{Weakly-supervised Methods}} \\

        \midrule

        MS-Former~\cite{li2023msformermemorysupportedtransformerweakly}
        & 71.69 & 82.85 & 76.87 & 62.43 & --
        & 86.63 & 79.13 & 82.71 & 70.52 & --
        & \textbf{86.34} & 61.99 & 72.17 & 56.45 & -- \\

        CARGNet~\cite{fang2023point}
        & 59.93 & \textbf{94.25} & 73.27 & 57.28 & 96.50
        & 44.47 & \textbf{92.20} & 60.00 & 42.86 & 95.12
        & 59.35 & \textbf{85.49} & 70.05 & 50.40 & 79.57 \\

        TransWCD~\cite{zhao2025transwcd}
        & 55.48 & 65.51 & 60.08 & 52.94 & 95.56
        & 75.34 & 65.19 & 68.73 & 52.36 & 97.17
        & -- & -- & -- & -- & -- \\

        BARNet~\cite{jiang2025barnet}
        & -- & -- & -- & -- & --
        & 68.70 & 70.70 & 69.45 & 53.20 & 94.53
        & -- & -- & -- & -- & -- \\

        PGU-CD~\cite{WANG2026101949}
        & 66.61 & 79.70 & 72.57 & 56.95 & 96.93
        & 88.04 & 76.43 & 81.82 & 69.24 & 98.57
        & -- & -- & -- & -- & -- \\

        \midrule

        \textbf{Ours}
        & \textbf{78.48} & 84.80 & \textbf{81.25} & \textbf{68.80} & \textbf{98.04}
        & \textbf{90.97} & 81.93 & \textbf{86.21} & \textbf{75.77} & \textbf{98.96}
        & 78.36 & 67.07 & \textbf{72.28} & \textbf{56.59} & \textbf{87.86} \\

        \bottomrule
    \end{tabular}
    }

    \caption{Quantitative comparison on three change detection benchmarks, where the best results are highlighted in bold.}
    \label{tab:comparison_main}
\end{table*}

\subsection{Experimental Setup.}

\noindent \textbf{Datasets.}
To validate the effectiveness of our framework, we conduct experiments on widely used high-resolution remote sensing change detection datasets: WHU-CD, LEVIR-CD, SYSU-CD. All image pairs are uniformly cropped into patches of 256×256 pixels. 
All datasets are split into training and testing sets following their official protocols to ensure fair comparisons with existing methods.
The point annotations are generated by sampling the centroids of connected components in the corresponding ground-truth change masks.

\noindent \textbf{Evaluation Metrics.} 
We adopt five widely used metrics to comprehensively evaluate the performance of different methods, including F1-score (\(F1\)), Intersection over Union (\(IoU\)), Precision (\(P\)), Recall (\(R\)), and Overall Accuracy (\(OA\)).

\noindent \textbf{Implementation Details.}
All experiments are conducted on an NVIDIA RTX A6000 GPU. 
In Stage I, the CNN refinement model is optimized using stochastic gradient descent (SGD) with an initial learning rate of \(1\times10^{-2}\), a momentum of 0.9, and a weight decay of \(5\times10^{-4}\). The model is trained for 8,000 iterations. 
In Stage II, we construct the teacher-student self-training framework. 
The student network is optimized using AdamW with an initial learning rate of \(5\times10^{-4}\), a weight decay of \(5\times10^{-4}\), a batch size of 8, and a total of 8 epochs. 
The input images are resized to \(256\times256\). A cosine annealing scheduler is employed to gradually update the learning rate. 
The teacher network is updated by exponential moving average (EMA) with a decay factor of 0.996.
In addition, the pseudo-labels are refreshed every 2 epochs, and the refresh momentum coefficient is set to 0.8.

\subsection{Compare with State-of-the-arts.}
To verify the effectiveness of the proposed framework, we compare it with 14 state-of-the-art methods on three benchmark datasets, including 5 weakly supervised and 9 fully supervised methods. For fair comparison, all competing methods are implemented and evaluated under their officially recommended settings.

\subsubsection{Quantitative analysis.}
From table~\ref{tab:comparison_main}, it can be observed that our method achieves the best overall performance among weakly supervised approaches on most datasets, demonstrating the effectiveness of introducing SAM priors and the proposed two-stage closed-loop optimization framework. 
It is worth noting that, although our method is trained under weak supervision, its performance is already close to, or even better than, some fully supervised methods on several datasets. 

\noindent 1) \textbf{Results on the LEVIR-CD:}
our method achieves 81.25\% F1 and 68.80\% IoU. 
Compared with fully supervised methods, the average performance drop is only 9.18\% in F1 and 12.73\% in IoU. 
Moreover, our method surpasses the fully supervised FC-Siam-Conc and FC-Siam-Diff in Recall, achieving 84.80\% versus 84.59\% and 81.65\%, respectively. 
Compared with the second-best weakly supervised method, MS-Former, our method improves F1 and IoU by 4.38 and 6.37 points, respectively.

\noindent 2) \textbf{Results on the WHU-CD:}
WHU-CD involves more complex and diverse change objects, our method still achieves 86.21\% F1 and 75.77\% IoU.
In this challenging setting, our method outperforms several fully supervised baselines, including FC-Siam-Conc, FC-Siam-Diff, and ChangeFormer. 
Compared with the best competing weakly supervised method, MS-Former, our method improves F1 and IoU by 3.50 and 5.25 points, respectively.

\noindent 3) \textbf{Results on the SYSU-CD:}
Our method achieves 72.28\% F1 and 56.59\% IoU on SYSU-CD. 
Compared with MS-Former, the strongest weakly supervised baseline, our method still achieves slightly better performance, with gains of 0.11\% in F1 and 0.14\% in IoU. Although the margin is not large, this result is still meaningful because SYSU-CD contains more complex scene variations, making weakly supervised pseudo-label learning more difficult.

\begin{table*}[t]
    \centering
    \footnotesize
    \setlength{\tabcolsep}{10pt}
    \renewcommand{\arraystretch}{1.1}
    \begin{tabular}{lcccccc}
        \toprule
        \multirow{2}{*}{pseudo-labels Used in Stage II}
        & \multicolumn{2}{c}{LEVIR-CD}
        & \multicolumn{2}{c}{WHU-CD}
        & \multicolumn{2}{c}{SYSU-CD} \\
        \cmidrule(lr){2-3}
        \cmidrule(lr){4-5}
        \cmidrule(lr){6-7}
        & $F1$ & $IoU$ & $F1$ & $IoU$ & $F1$ & $IoU$ \\
        \midrule
        Initial SAM2 pseudo-labels                & 72.22 & 56.52 & 83.12 & 71.12 & 52.38 & 35.48 \\
        Refined pseudo-labels from Stage I        & 79.42 & 65.62 & 84.91 & 73.79 & 67.39 & 50.79 \\
        Initial SAM2 pseudo-labels + Self-Training & 79.68 & 66.23 & 83.70 & 72.08 & 64.92 & 48.06 \\
        \midrule
        Full Model                                & \textbf{81.25} & \textbf{68.80} & \textbf{86.21} & \textbf{75.77} & \textbf{72.28} & \textbf{56.59} \\
        \bottomrule
    \end{tabular}
    \caption{Ablation study on how different settings affect the second-stage training across datasets.}
    \label{tab:ablation_stages_multi}
\end{table*}

\subsubsection{Qualitative analysis.}

Figures~\ref{fig:vis-Compare} show qualitative comparisons with both fully supervised and weakly supervised methods on benchmark datasets. 
Compared with existing methods, the proposed approach produces more complete change regions, clearer boundaries, and fewer false alarms. 
This advantage is particularly evident in small structures and complex scenes, where our method better preserves structural details while suppressing background noise. These observations are consistent with the quantitative results and show that the proposed method remains competitive with both weakly supervised and several fully supervised approaches.

\begin{table}[t]
    \centering
    \scriptsize
    \setlength{\tabcolsep}{5pt}
    \renewcommand{\arraystretch}{1.1}
    \begin{tabular}{lccccc}
        \toprule
        Method Variant & $P$ & $R$ & $F1$ & $IoU$ & $OA$ \\
        \midrule
        w/o Bi-Temporal Mask Selection & 77.70 & 74.74 & 76.19 & 61.54 & 97.62 \\
        w/o Uncertainty-Aware Loss     & 60.14 & 71.43 & 65.30 & 48.48 & 96.13 \\
        w/o Teacher EMA                & 77.20 & 79.49 & 78.32 & 64.38 & 97.76 \\
        \midrule
        Full Model                     & \textbf{78.48} & \textbf{84.80} & \textbf{81.25} & \textbf{68.80} & \textbf{98.04} \\
        \bottomrule
    \end{tabular}
    \caption{Ablation study of key components in Stage I and Stage II on the LEVIR-CD dataset.}
    \label{tab:ablation_modules}
\end{table}

\subsection{Ablation Study.}

To further verify the effectiveness of each component in the proposed framework, we conduct ablation experiments from three aspects: 
1) the contribution of each stage, 
2) the effectiveness of key modules in Stage I and Stage II, and 
3) the sensitivity of important hyper-parameters. Unless otherwise specified, all ablation experiments are conducted on the LEVIR-CD dataset under the same settings as the full model.

\def\w{0.95\linewidth}
\def\h{1.0in}
\begin{figure}[tbp]
    \setlength{\tabcolsep}{1.0pt}
    \centering
    \small
    \begin{tabular}{@{}c@{}}
        \includegraphics[width=\w]{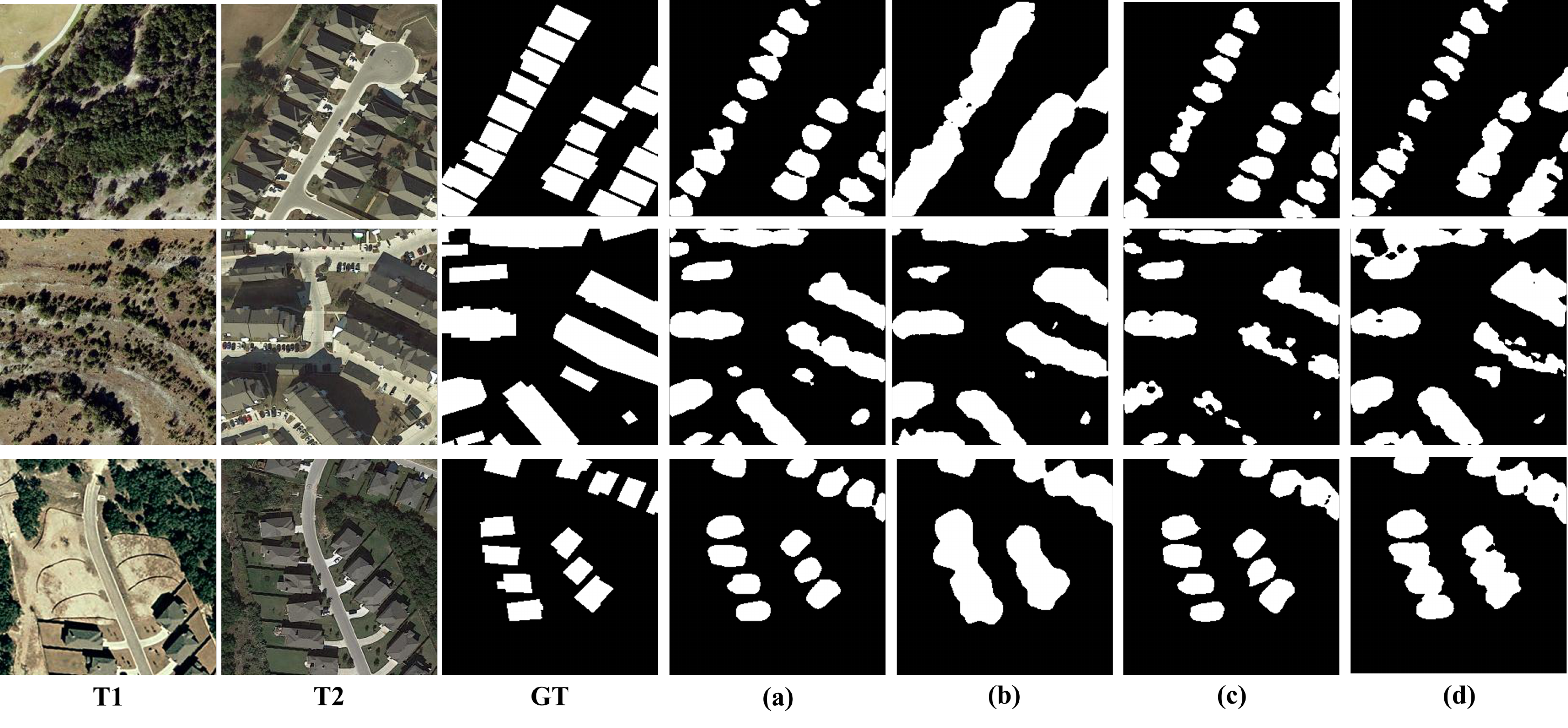}
    \end{tabular}
    \caption{Qualitative ablation results. (a) Full model, (b) initial SAM pseudo-labels only, (c) initial pseudo-labels with CNN refinement (Stage I), and (d) initial pseudo-labels with self-training (Stage II).}
    \label{fig:Abli}
\end{figure}
\begin{table}[t]
    \centering
    \scriptsize
    \setlength{\tabcolsep}{5pt}
    \renewcommand{\arraystretch}{1.08}
    \begin{tabular}{lcccccc}
        \toprule
        \multirow{2}{*}{Dataset}
        & \multicolumn{3}{c}{F1@0.5 $\uparrow$}
        & \multicolumn{3}{c}{MAE $\downarrow$} \\
        \cmidrule(lr){2-4}
        \cmidrule(lr){5-7}
        & Initial & Stage I & Stage II
        & Initial & Stage I & Stage II \\
        \midrule
        LEVIR-CD & 29.87 & 32.49 & \textbf{33.36}
                 & 0.036 & 0.024 & \textbf{0.022} \\
        WHU-CD   & 17.82 & 19.74 & \textbf{20.04}
                 & 0.032 & 0.018 & \textbf{0.017} \\
        SYSU-CD  & 50.23 & 58.08 & \textbf{58.19}
                 & 0.164 & 0.147 & \textbf{0.143} \\
        \bottomrule
    \end{tabular}
    \caption{Pseudo-label quality across optimization stages.}
    \label{tab:pseudo_label_quality}
\end{table}

\textbf{Impact of Each Stage.}
Table~\ref{tab:ablation_stages_multi} reports the contribution of different pseudo-label settings on LEVIR-CD, WHU-CD, and SYSU-CD. Using only the initial SAM2 pseudo-labels provides a reasonable baseline under weak supervision, but the results are still limited by noisy and incomplete supervision. Introducing the CNN refinement module in Stage I consistently improves performance on all three datasets, showing its effectiveness in enhancing pseudo-label quality. Adding self-training in Stage II also brings clear gains over the initial pseudo-label baseline, validating the benefit of iterative pseudo-label optimization. The full model achieves the best results on all three datasets, demonstrating that Stage I refinement and Stage II self-training are complementary.

\textbf{Pseudo-Label Quality Analysis.}
Table~\ref{tab:pseudo_label_quality} directly evaluates pseudo-label quality against ground-truth masks. Across all three datasets, the pseudo-labels are progressively improved from the initial SAM2 outputs to Stage I refinement and Stage II refresh, as reflected by higher F1@0.5 and lower MAE. This confirms that the proposed optimization process improves the supervision signal itself rather than only the final detector.

\textbf{Impact of key modules in Stage I and Stage II.}
There are several key components in our framework:
(i) Mask selection in Stage I. A pivotal part of Stage I is the bi-temporal mask selection strategy, which ranks and fuses SAM-generated candidate masks by jointly considering SAM confidence, change consistency, and point coverage. As shown in Table~\ref{tab:ablation_modules}, removing this strategy leads to consistent performance drops, validating its effectiveness in generating more reliable initial pseudo-labels.
(ii) Uncertainty-aware loss in Stage I. We further evaluate the role of the uncertainty-aware weighting scheme by replacing it with a standard loss without confidence weighting. The performance decline indicates that uncertainty-aware supervision can better suppress noisy pseudo-labels and guide the model to focus on more reliable regions.
(iii) Teacher-student self-training in Stage II. To validate the role of the EMA update strategy, we replace the EMA-updated teacher with a non-EMA variant while keeping the rest of the self-training framework unchanged. The inferior results indicate that EMA updating provides more stable supervision and is important for progressively improving pseudo-label quality and task-specific adaptation.

\begin{table}[tp]
    \centering
    \footnotesize
    \setlength{\tabcolsep}{4pt}
    \renewcommand{\arraystretch}{1.1}
    \begin{tabular}{c|ccccc}
        \hline
        Refresh interval & $P$ & $R$ & $F1$ & $IoU$ & $OA$ \\
        \hline
        1 epoch        & 77.71 & 83.43 & 79.91 & 66.54 & 97.93 \\
        2 epochs (Ours) & \textbf{78.48} & \textbf{84.80} & \textbf{81.25} & \textbf{68.80} & \textbf{98.04} \\
        3 epochs       & 79.77 & 81.10 & 80.77 & 67.27 & 97.99 \\
        4 epochs       & 79.21 & 82.39 & 80.43 & 67.74 & 98.01 \\
        \hline
    \end{tabular}
    \caption{Sensitivity analysis of the pseudo-label refresh interval on the LEVIR-CD dataset.}
    \label{tab:ablation_interval}
\end{table}

\textbf{Impact of the Pseudo-Label Refresh Interval.}
Table~\ref{tab:ablation_interval} shows the sensitivity of the pseudo-label refresh interval in Stage II. 
The best performance is achieved when the pseudo-labels are refreshed every 2 epochs, obtaining 81.25\% F1 and 68.80\% IoU. 
Refreshing too frequently degrades the performance, since the pseudo-labels may become unstable before the student network is sufficiently optimized. 
In contrast, overly sparse refreshing also hurts performance because outdated pseudo-labels limit the benefit of self-training. 
These results suggest that a moderate refresh interval provides the best balance between stability and adaptivity.

\section{Conclusion}

In this paper, we propose a two-stage point-supervised change detection framework that exploits foundation-model priors and iterative pseudo-label optimization. 
In Stage I, SAM2 is used to generate candidate masks from bi-temporal point supervision, and a bi-temporal mask selection strategy together with CNN-based refinement is introduced to construct higher-quality soft pseudo-labels. 
In Stage II, a teacher-student self-training framework is further developed to progressively refresh pseudo-labels and improve task-specific adaptation. 
We also implement the Stage I candidate generation and bi-temporal selection mechanism as a point-interactive change annotation tool.
Extensive experiments on three benchmark datasets show that the proposed method achieves the best results among the compared weakly supervised approaches, while remaining competitive with several fully supervised baselines.
In future work, we will further study how to improve pseudo-label reliability on challenging scenes and extend the proposed framework to more general forms of weak supervision.

\section*{LLM Usage Statement}
We used ChatGPT (OpenAI) to assist with English language polishing and grammar correction in the manuscript, and OpenAI Codex to assist with code generation for the implementation. These tools were used in an assistive capacity; the authors remain fully responsible for the submitted text, code, experimental results, and scientific claims.

{
    \small
    \bibliographystyle{ieeenat_fullname}
    \bibliography{main}

\begin{thebibliography}{26}
\providecommand{\natexlab}[1]{#1}
\providecommand{\url}[1]{\texttt{#1}}
\expandafter\ifx\csname urlstyle\endcsname\relax
  \providecommand{\doi}[1]{doi: #1}\else
  \providecommand{\doi}{doi: \begingroup \urlstyle{rm}\Url}\fi

\bibitem[Bandara(2022)]{9883686}
Wele Gedara~Chaminda Bandara.
\newblock A transformer-based siamese network for change detection.
\newblock In \emph{IGARSS 2022 - 2022 IEEE International Geoscience and Remote
  Sensing Symposium}, pages 207--210, 2022.

\bibitem[Caye~Daudt et~al.(2018)Caye~Daudt, Le~Saux, and Boulch]{8451652}
Rodrigo Caye~Daudt, Bertr Le~Saux, and Alexandre Boulch.
\newblock Fully convolutional siamese networks for change detection.
\newblock In \emph{2018 25th IEEE International Conference on Image Processing
  (ICIP)}, pages 4063--4067, 2018.

\bibitem[Ding et~al.(2024)Ding, Zhu, Peng, Tang, Yang, and Bruzzone]{Ding_2024}
Lei Ding, Kun Zhu, Daifeng Peng, Hao Tang, Kuiwu Yang, and Lorenzo Bruzzone.
\newblock Adapting segment anything model for change detection in vhr remote
  sensing images.
\newblock \emph{IEEE Transactions on Geoscience and Remote Sensing},
  62:\penalty0 1–11, 2024.

\bibitem[Dong et~al.(2024)Dong, Wang, Du, and Meng]{DONG202453}
Sijun Dong, Libo Wang, Bo Du, and Xiaoliang Meng.
\newblock Changeclip: Remote sensing change detection with multimodal
  vision-language representation learning.
\newblock \emph{ISPRS Journal of Photogrammetry and Remote Sensing},
  208:\penalty0 53--69, 2024.

\bibitem[Fang et~al.(2023)Fang, Jiang, Yu, Zhang, and Yue]{fang2023point}
Leyuan Fang, Yiqi Jiang, Hongfeng Yu, Yingying Zhang, and Jun Yue.
\newblock Point label meets remote sensing change detection: A
  consistency-aligned regional growth network.
\newblock \emph{IEEE Transactions on Geoscience and Remote Sensing}, 2023.

\bibitem[{Fang} et~al.(2021){Fang}, {Li}, {Shao}, and {Li}]{9355573}
S. {Fang}, K. {Li}, J. {Shao}, and Z. {Li}.
\newblock Snunet-cd: A densely connected siamese network for change detection
  of vhr images.
\newblock \emph{IEEE Geoscience and Remote Sensing Letters}, pages 1--5, 2021.

\bibitem[Han(2023)]{han2023hcgmnethierarchicalchangeguiding}
Chengxi Han.
\newblock Hcgmnet: A hierarchical change guiding map network for change
  detection, 2023.

\bibitem[Han et~al.(2023)Han, Wu, Guo, Hu, Li, and Chen]{10093022}
Chengxi Han, Chen Wu, Haonan Guo, Meiqi Hu, Jiepan Li, and Hongruixuan Chen.
\newblock Change guiding network: Incorporating change prior to guide change
  detection in remote sensing imagery.
\newblock \emph{IEEE Journal of Selected Topics in Applied Earth Observations
  and Remote Sensing}, pages 1--17, 2023.

\bibitem[Hao~Chen and Shi(2021)]{chen2021a}
Zipeng~Qi Hao~Chen and Zhenwei Shi.
\newblock Remote sensing image change detection with transformers.
\newblock \emph{IEEE Transactions on Geoscience and Remote Sensing}, pages
  1--14, 2021.

\bibitem[He et~al.(2025)He, Zhang, Tang, Yang, Li, Fan, and
  Farsiu]{he2025scalersamenhancedcollaborativelearning}
Chunming He, Rihan Zhang, Longxiang Tang, Ziyun Yang, Kai Li, Deng-Ping Fan,
  and Sina Farsiu.
\newblock Scaler: Sam-enhanced collaborative learning for label-deficient
  concealed object segmentation, 2025.

\bibitem[Jiang et~al.(2025)Jiang, Zhong, Zhang, Gong, Zhou, Zhao, and
  Guan]{jiang2025barnet}
Fenlong Jiang, Zikang Zhong, Mingyang Zhang, Maoguo Gong, Yu Zhou, Wei Zhao,
  and Ziyu Guan.
\newblock Barnet: Boundary-aware refinement network for weakly supervised
  change detection.
\newblock \emph{IEEE Transactions on Geoscience and Remote Sensing}, 2025.

\bibitem[Kage et~al.(2026)Kage, Rothenberger, Andreadis, and
  Diochnos]{Kage_2026}
Patrick Kage, Jay Rothenberger, Pavlos Andreadis, and Dimitrios Diochnos.
\newblock A review of pseudo-labeling for computer vision.
\newblock \emph{Journal of Artificial Intelligence Research}, 85, 2026.

\bibitem[Kirillov et~al.(2023)Kirillov, Mintun, Ravi, Mao, Rolland, Gustafson,
  Xiao, Whitehead, Berg, Lo, Dollár, and Girshick]{kirillov2023segment}
Alexander Kirillov, Eric Mintun, Nikhila Ravi, Hanzi Mao, Chloe Rolland, Laura
  Gustafson, Tete Xiao, Spencer Whitehead, Alexander~C. Berg, Wan-Yen Lo, Piotr
  Dollár, and Ross Girshick.
\newblock Segment anything, 2023.

\bibitem[Lei et~al.(2024)Lei, Xu, Ning, Lv, Min, Jin, and Nandi]{10286068}
Tao Lei, Yetong Xu, Hailong Ning, Zhiyong Lv, Chongdan Min, Yaochu Jin, and
  Asoke~K. Nandi.
\newblock Lightweight structure-aware transformer network for remote sensing
  image change detection.
\newblock \emph{IEEE Geoscience and Remote Sensing Letters}, 21:\penalty0 1--5,
  2024.

\bibitem[Li et~al.(2023)Li, Tang, Liu, Li, Li, and
  Zhang]{li2023msformermemorysupportedtransformerweakly}
Zhenglai Li, Chang Tang, Xinwang Liu, Changdong Li, Xianju Li, and Wei Zhang.
\newblock Ms-former: Memory-supported transformer for weakly supervised change
  detection with patch-level annotations, 2023.

\bibitem[Liu et~al.(2025{\natexlab{a}})Liu, Xu, Su, Zhang, and Li]{10839471}
Nanqing Liu, Xun Xu, Yongyi Su, Haojie Zhang, and Heng-Chao Li.
\newblock Pointsam: Pointly-supervised segment anything model for remote
  sensing images.
\newblock \emph{IEEE Transactions on Geoscience and Remote Sensing},
  63:\penalty0 1--15, 2025{\natexlab{a}}.

\bibitem[Liu et~al.(2025{\natexlab{b}})Liu, Xu, Lei, Wang, Du, Zhang, Lv, and
  Gong]{AEKAN}
Tongfei Liu, Jianjian Xu, Tao Lei, Yingbo Wang, Xiaogang Du, Weichuan Zhang,
  Zhiyong Lv, and Maoguo Gong.
\newblock Aekan: Exploring superpixel-based autoencoder kolmogorov-arnold
  network for unsupervised multimodal change detection.
\newblock \emph{IEEE Transactions on Geoscience and Remote Sensing},
  63:\penalty0 1--14, 2025{\natexlab{b}}.

\bibitem[Meyer et~al.(2026)Meyer, Mutter, and
  Padoy]{meyer2026dexterweaklysemisupervisedobject}
Adrien Meyer, Didier Mutter, and Nicolas Padoy.
\newblock Dexter: Weakly semi-supervised object detection with class and
  instance experts for medical imaging, 2026.

\bibitem[Ning et~al.(2025)Ning, He, Lei, Cao, Zhang, Chen, and Nandi]{11193785}
Hailong Ning, Qi He, Tao Lei, Xiaopeng Cao, Wuxia Zhang, Yanping Chen, and
  Asoke~K. Nandi.
\newblock Da2-net: Integrating sam2 with domain adaption and difference
  aggregation for remote sensing change detection.
\newblock \emph{IEEE Transactions on Geoscience and Remote Sensing},
  63:\penalty0 1--17, 2025.

\bibitem[Ravi et~al.(2024)Ravi, Gabeur, Hu, Hu, Ryali, Ma, Khedr, R{\"a}dle,
  Rolland, Gustafson, Mintun, Pan, Alwala, Carion, Wu, Girshick, Doll{\'a}r,
  and Feichtenhofer]{ravi2024sam2}
Nikhila Ravi, Valentin Gabeur, Yuan-Ting Hu, Ronghang Hu, Chaitanya Ryali,
  Tengyu Ma, Haitham Khedr, Roman R{\"a}dle, Chloe Rolland, Laura Gustafson,
  Eric Mintun, Junting Pan, Kalyan~Vasudev Alwala, Nicolas Carion, Chao-Yuan
  Wu, Ross Girshick, Piotr Doll{\'a}r, and Christoph Feichtenhofer.
\newblock Sam 2: Segment anything in images and videos.
\newblock \emph{arXiv preprint arXiv:2408.00714}, 2024.

\bibitem[Subhani(2026)]{subhani2026resam}
Muhammad~Naseer Subhani.
\newblock Resam: Refine, requery, and reinforce: Self-prompting
  point-supervised segmentation for remote sensing images.
\newblock In \emph{Proceedings of the IEEE/CVF Conference on Computer Vision
  and Pattern Recognition (CVPR)}, 2026.

\bibitem[Wang et~al.(2023)Wang, Zhang, and Shi]{10310006}
Lukang Wang, Min Zhang, and Wenzhong Shi.
\newblock Cs-wscdnet: Class activation mapping and segment anything model-based
  framework for weakly supervised change detection.
\newblock \emph{IEEE Transactions on Geoscience and Remote Sensing},
  61:\penalty0 1--12, 2023.

\bibitem[Wang et~al.(2026)Wang, Li, Samat, Liu, and Li]{WANG2026101949}
Yongqing Wang, Erzhu Li, Alim Samat, Wei Liu, and Xing Li.
\newblock Pgu-cd: A point-guided uncertainty-aware framework for building
  change detection.
\newblock \emph{Remote Sensing Applications: Society and Environment}, page
  101949, 2026.

\bibitem[Wei(2025)]{rs17030369}
Chenlong Wei.
\newblock Ass-cd: Adapting segment anything model and swin-transformer for
  change detection in remote sensing images.
\newblock \emph{Remote Sensing}, 17\penalty0 (3), 2025.

\bibitem[Wu(2022)]{wu2022fullyconvolutionalchangedetection}
Chen Wu.
\newblock Fully convolutional change detection framework with generative
  adversarial network for unsupervised, weakly supervised and regional
  supervised change detection, 2022.

\bibitem[Zhao et~al.(2025)Zhao, Ru, Wu, and Wang]{zhao2025transwcd}
Zhenghui Zhao, Lixiang Ru, Chen Wu, and Di Wang.
\newblock Transwcd: Scene-adaptive joint constrained framework for weakly
  supervised change detection.
\newblock \emph{IEEE Transactions on Geoscience and Remote Sensing},
  63:\penalty0 1--12, 2025.

\end{thebibliography}
}
\end{document}